\documentstyle[11pt]{article}
\title{Spanish tagset for the CRATER project}
\author{Fernando S\'anchez Le\'on \\
Laboratorio de Ling\"u\'{\i}stica Inform\'atica \\
Facultad de Filosof\'{\i}a y Letras \\
Universidad Aut\'onoma de Madrid \\
e-mail: fsanchez@ccuam3.sdi.uam.es}
\date{March 7, 1994}

\begin{document}
\maketitle

\section{Introduction}

This document contains the second version of the Spanish tagset for the
CRATER project. It has been extensively changed with respect to the
first draft of the tagset, dated January 28\footnote{I am indebted to
Flora Ram\'{\i}rez Bustamante, who has taken active part in the
refinement of the first draft of this document.}. The major motivations
for changes are pointed out in section 2. For a more comprehensive
exposition of the changes, refer to \cite{FSLcomp_ts}. Section 3
contains the tagset with an English description of each tag. \\

\section{Design motivations}

In its origins, this tagset was inspired by the English tagset used in
the ET10/63 project, which, in turn, was derived from the one used by
CLAWS \cite{TCAE}.  The kind of information tackled by the English
tagset was conveniently adapted to Spanish.  Major morphosyntactic
categories were contemplated, but also some syntactic and semantic
distinctions beyond the level of morphosyntax were taken into account.
For instance, semantic classes for temporal, measure or locative nouns
have their own tags.  The naming conventions of the English tagset,
though, were modified as to produce mnemonic tags, as far as possible.
\\

The first version contained a set of errors and inconsistencies that
needed a correction. Other decisions had to be reviewed on the light of
two major issues:  the recommendations of both TEI and EAGLES on text
annotation and the idea that the tagset was to be used by an automatic
tagging system.  \\

The objectives of CRATER include the development of a public domain POS
tagger for Spanish and the production of a (sample) tagged corpus of over
one million words. This goal will contribute to the creation of valuable
corpus and linguistic resources for Spanish.

One of the major concerns of the computational linguistics community in
the last years has been the reusability of resources of varied types.
Consequently, in order to define the characteristics for future
resources to be reusable, a set of initiatives and working groups has
been set up to produce guidelines and recommendations aiming at the
standardization of the type and form of the information to be included
in these linguistic resources.

One of these groups is the Text Encoding Initiative. TEI produced a
first draft of its Guidelines \cite{TEIP1} in 1990 and, since
then, has delivered a set of draft documents related to the annotation
of texts \cite{AI1W3}, \cite{AI1W9}, \cite{AI1W2}, that will
conform the relevant chapters of the second version of its guidelines,
known as TEI P2.

Other working groups, like EAGLES ({\it Expert Advisory Groups on
Language Engineering Standards}), are also working on its own
recommendations for morphosyntactic annotation.  For the moment, only a
draft document for sections 4.6 and 4.7 has been produced \cite{MSAL21}.

Therefore, if linguistic resources for CRATER are to be considered
reusable, it has to be guaranteed that they follow these
recommendations ---at least concerning the type of information to be
included in the annotations, since the {\it syntax} can be produced with
the appropriate mappings. This is particularly important if we consider
that CRATER will create a POS tagger, which can potentially produce new
resources for Spanish.

Thus, the tagset has been extended as to include (almost) all the
feature specificacions proposed by TEI and/or EAGLES. \\

The other issue goes, somehow, in the opposite direction. If the goal
is to produce an accurate tagged corpus by automatic means and taking
into account only morphosyntax, then, the tagset must be devised in a
way that allows the system to assign a given tag to a word in the
corpus with a fair guarantee of success. This is particularly important
in the case of open classes not included in the lexicon. Thus, former
tags used to stress lexical properties of nouns like their ability to
be masculine or feminine depending on their referent ({\it amigo,
amiga}, friend) have been eliminated. They would be impossible to
assign if not explicitly included in the lexicon. Besides, the
distinction lies more in a lexicon view of morphosyntactic annotation
\cite[p. 10]{MSAL21}. Other distinctions not treated already in
the previous version of the tagset include homographs which could only
be disambiguated taken into account a deep syntactic and/or semantic
analysis of the structures they appear in. Some forms of the numeral
{\it uno} (one) and the indefinite article are one the cases of
homography collapsed in one tag. The pronoun {\it se} is other
example.

Apart from these minor changes in the direction of restricting the
tagset, no other limiting measures have been taken, although it is a
common practice for projects using automatic tagging systems to pair
down as much as possible the number of tags \cite{PENN92}. On the
contrary, the richness and wideness of the information considered has
the consequence of producing various ambiguity classes. Nevertheless,
these will be properly accounted for within the automatic tagging
system with the aid of {\it transition biases}, a set of automata for
validating or forbidding specific sequences of tags. \\

There is still one issue to be addressed ---the correspondence between
textwords and orthographic words as segmented by the tokenizer program.
Nevertheless, this is not a problem that affects the tagset, since it
should be possible, in principle, to tag textwords occurring in complex
ortographic words with the corresponding tags or assign the correct tag
to multiple ortographic words representing only one textword.  The
first mapping problem ---the correspondence between more than one
textword and one ortographic word--- appears in portmanteau words and
verbs with enclitic forms. While the former are included in the tagset
with a tag on their own, there are no specific tags for the latter,
since the number of verb and enclitic(s) combinations makes prohibitive
the definition of special tags for them. This entails a certain
inconsistency in their treatment, but the solution to be adopted, which
depends primarily on the tagger, should be finally consistent.

The second problem ---the correspondence between one textword and more
than one ortographic word--- will need also a solution that, again,
depends on the software.

\section{List of grammatical tags for Spanish}

\begin{description}

\item[IQUEST] Punctuation tag - question mark (inverted)
\item[IEXCL ] Punctuation tag - exclamation mark (inverted)
\item[!     ] Punctuation tag - exclamation mark
\item["     ] Punctuation tag - quotes
\item[(     ] Punctuation tag - left bracket
\item[)     ] Punctuation tag - right bracket
\item[,     ] Punctuation tag - comma
\item[-     ] Punctuation tag - dash
\item[.     ] Punctuation tag - full-stop
\item[...   ] Punctuation tag - ellipsis
\item[:     ] Punctuation tag - colon
\item[;     ] Punctuation tag - semicolon
\item[?     ] Punctuation tag - question mark
\item[ADJCP ] Plural general comparative adjective (mayores, menores)
\item[ADJCS ] Singular general comparative adjective (mayor, menor)
\item[ADJGFP] Feminine plural general positive adjective
\item[ADJGFS] Feminine singular general positive adjective
\item[ADJGMP] Masculine plural general positive adjective
\item[ADJGMS] Masculine singular general positive adjective
\item[ADJSFP] Feminine plural general superlative adjective (m\'aximas,
m\'{\i}nimas)
\item[ADJSFS] Feminine singular general superlative adjective (m\'axima,
m\'{\i}nima)
\item[ADJSMP] Masculine plural general superlative adjective (m\'aximos,
m\'{\i}nimos)
\item[ADJSMS] Masculine singular general superlative adjective (m\'aximo,
m\'{\i}nimo, grand\'{\i}simo)
\item[ADVGR ] Positive degree adverb (muy, demasiado, mucho)
\item[ADVGRC] Comparative degree adverb (m\'as, menos)
\item[ADVGRS] Superlative degree adverb (abundant\'{\i}simamente)
\item[ADVINT] Interrogative adverb (c\'omo)
\item[ADVL  ] Locative adverb underspecified for directionality (abajo)
\item[ADVLD ] Dynamic locative adverb (adelante)
\item[ADVLE ] Static locative adverb (dentro)
\item[ADVLIN] Interrogative locative adverb (d\'onde)
\item[ADVLP ] Locative adverb with proximal deixis (aqu\'{\i})
\item[ADVLR ] Locative adverb with remote deixis (all\'{\i})
\item[ADVLRD] Relative dynamic locative adverb (adonde)
\item[ADVLRE] Relative locative adverb underspecified for directionality
(donde)
\item[ADVN  ] General adverb (salvajemente, bien, probablemente)
\item[ADVNEG] General negative adverb (tampoco)
\item[ADVMRE] Relative modal adverb (como)
\item[ADVT  ] Temporal adverb (ahora, ayer)
\item[ADVTIN] Interrogative temporal adverb (cu\'ando)
\item[ADVTNE] Negative temporal adverb (nunca)
\item[ADVTRE] Relative temporal adverb (cuando)
\item[ALFP  ] Plural letter of the alphabet (As/Aes, bes)
\item[ALFS  ] Singular letter of the alphabet (A, b)
\item[ARCAFS] Feminine singular indefinite article and cardinal capable of
pronominal function (una)
\item[ARCAMS] Masculine singular indefinite article and non pronominal cardinal
(un)
\item[ARTDFP] Feminine plural definite article (las)
\item[ARTDFS] Feminine singular definite article (la)
\item[ARTDMP] Masculine plural definite article (los)
\item[ARTDMS] Masculine singular definite article (el)
\item[ARTDNS] Neuter singular definite article (lo)
\item[ARQUFP] Feminine plural indefinite article and quantifier capable of
pronominal function (unas)
\item[ARQUMP] Masculine plural indefinite article and quantifier capable of
pronominal function (unos)
\item[CARDGU] Hyphenated cardinals (40-50, 1850-1990)
\item[CARDFP] Plural femenine cardinal capable of pronominal function
(doscientas)
\item[CARDMP] Plural masculine cardinal capable of pronominal function
(doscientos)
\item[CARDPS] Singular pronominal cardinal (uno)
\item[CARDXP] Plural cardinal neutral for gender (dos, tres, mil)
\item[CARNMP] Non pronominal plural masculine cardinal (veinti\'un)
\item[CC    ] Coordinating conjunction (y, o)
\item[CCAD  ] Adversative coordinating conjunction (pero)
\item[CCNEG ] Negative coordinating conjunction (ni)
\item[CODE  ] Alphanumeric code
\item[CQUE  ] {\it que} (as conjunction)
\item[CSUBF ] Subordinating conjunction that introduces finite clauses (apenas)
\item[CSUBI ] Subordinating conjunction that introduces infinite clauses (al)
\item[CSUBX ] Subordinating conjunction underspecified for {\it subord-type}
(aunque)
\item[DMDPFP] Pronominal feminine plural demonstrative with distal deixis
(\'esas)
\item[DMDPFS] Pronominal feminine singular demonstrative with distal deixis
(\'esa)
\item[DMDPMP] Pronominal masculine plural demonstrative with distal deixis
(\'esos)
\item[DMDPMS] Pronominal masculine singular demonstrative with distal deixis
(\'ese)
\item[DMDPNS] Pronominal neuter singular demonstrative with distal deixis (eso)
\item[DMDXFP] Feminine plural demonstrative (capable of pronominal function)
with distal deixis (esas)
\item[DMDXFS] Feminine singular demonstrative (capable of pronominal function)
with distal deixis (esa)
\item[DMDXMP] Masculine plural demonstrative (capable of pronominal function)
with distal deixis (esos)
\item[DMDXMS] Masculine singular demonstrative (capable of pronominal function)
with distal deixis (ese)
\item[DMPPFP] Pronominal feminine plural demonstrative with proximal deixis
(\'estas)
\item[DMPPFS] Pronominal feminine singular demonstrative with proximal deixis
(\'esta)
\item[DMPPMP] Pronominal masculine plural demonstrative with proximal deixis
(\'estos)
\item[DMPPMS] Pronominal masculine singular demonstrative with proximal deixis
(\'este)
\item[DMPXFP] Feminine plural demonstrative (capable of pronominal function)
with proximal deixis (estas)
\item[DMPXFS] Feminine singular demonstrative (capable of pronominal function)
with proximal deixis (esta)
\item[DMPXMP] Masculine plural demonstrative (capable of pronominal function)
with proximal deixis (estos)
\item[DMPXMS] Masculine singular demonstrative (capable of pronominal function)
with proximal deixis (este)
\item[DMRPFP] Pronominal feminine plural demonstrative with remote deixis
(aqu\'ellas)
\item[DMRPFS] Pronominal feminine singular demonstrative with remote deixis
(aqu\'ella)
\item[DMRPMP] Pronominal masculine plural demonstrative with remote deixis
(aqu\'ellos)
\item[DMRPMS] Pronominal masculine singular demonstrative with remote deixis
(aqu\'el)
\item[DMRPNS] Pronominal neuter singular demonstrative with remote deixis
(aquello)
\item[DMRXFP] Feminine plural demonstrative (capable of pronominal function)
with remote deixis (aquellas)
\item[DMRXFS] Feminine singular demonstrative (capable of pronominal function)
with remote deixis (aquella)
\item[DMRXMP] Masculine plural demonstrative (capable of pronominal function)
with remote deixis (aquellos)
\item[DMRXMS] Masculine singular demonstrative (capable of pronominal function)
with remote deixis (aquel)
\item[DMPPNS] Pronominal neuter singular demonstrative with proximal deixis
(esto)
\item[FO    ] Formula
\item[INTPXP] Plural interrogative pronoun for animates neutral for gender
(qui\'enes)
\item[INTPXS] Singular interrogative pronoun for animates neutral for gender
(qui\'en)
\item[INTXXX] Interrogative capable of pronominal function neutral for gender
and number (qu\'e)
\item[INTXFP] Feminine plural interrogative capable of pronominal function
(cu\'antas)
\item[INTXFS] Feminine singular interrogative capable of pronominal function
for inanimates (cu\'anta)
\item[INTXMP] Masculine plural interrogative capable of pronominal function
(cu\'antos)
\item[INTXMS] Masculine and neuter singular interrogative capable of pronominal
function for inanimates (cu\'anto)
\item[INTXXP] Plural interrogative neutral for gender capable of pronominal
function (cu\'ales)
\item[INTXXS] Singular interrogative neutral for gender capable of pronominal
function (cu\'al)
\item[ITJN  ] Interjection (oh, ja)
\item[NCFP  ] Feminine plural common noun (mesas, manos)
\item[NCFS  ] Feminine singular common noun (mesa, mano)
\item[NCMP  ] Masculine plural common noun (libros, ordenadores)
\item[NCMS  ] Masculine singular common noun (libro, ordenador)
\item[NEG   ] Negation
\item[NLOCFP] Feminine plural locative noun (islas, avenidas)
\item[NLOCFS] Feminine singular locative noun (isla, calle)
\item[NLOCMP] Masculine plural locative noun (montes)
\item[NLOCMS] Masculine singular locative noun (monte)
\item[NMEAFP] Feminine plural unit of measure noun (hect\'areas, micras)
\item[NMEAFS] Feminine singular unit of measure noun (hect\'area, micra)
\item[NMEAMP] Masculine plural unit of measure noun (metros, litros)
\item[NMEAMS] Masculine singular unit of measure noun (metro, litro)
\item[NNUMFP] Feminine plural numeral noun (docenas)
\item[NNUMFS] Feminine singular numeral noun (docena)
\item[NNUMMP] Masculine plural numeral noun (millares)
\item[NNUMMS] Masculine singular numeral noun (millar, tercio)
\item[NORGFP] Feminine plural organization noun (confederaciones)
\item[NORGFS] Feminine singular organization noun (confederaci\'on)
\item[NORGMP] Masculine plural organization noun (gobiernos, comit\'es)
\item[NORGMS] Masculine singular organization noun (consejo, departamento)
\item[NPAFP ] Feminine plural proper anthroponymous noun (Mar\'{\i}as)
\item[NPAFS ] Feminine singular proper anthroponymous noun (Mar\'{\i}a)
\item[NPAMP ] Masculine plural proper anthroponymous noun (Juanes)
\item[NPAMS ] Masculine singular proper anthroponymous noun (Juan)
\item[NPAXX ] Proper anthroponymous noun neutral for gender and number
(Rodr\'{\i}guez, Sanch\'{\i}s)
\item[NPTOP ] Plural proper toponym or organization noun (Coreas)
\item[NPTOS ] Singular proper toponym or organization noun (IBM, Madrid)
\item[NPTP  ] Plural proper toponym noun (Pirineos)
\item[NPTS  ] Singular proper toponym noun (Guadalquivir)
\item[NTMPFP] Feminine plural temporal noun (semanas, quincenas)
\item[NTMPFS] Feminine singular temporal noun (semana, quincena)
\item[NTMPMP] Masculine plural temporal noun (d\'{\i}as, a\~nos)
\item[NTMPMS] Masculine singular temporal noun (d\'{\i}a, a\~no)
\item[ORDNMS] Masculine singular non pronominal ordinal (primer, tercer)
\item[ORDXFP] Feminine plural ordinal capable of pronominal function (primeras,
segundas)
\item[ORDXFS] Feminine singular ordinal capable of pronominal function
(primera, segunda)
\item[ORDXMP] Masculine plural ordinal capable of pronominal function
(primeros, segundos)
\item[ORDXMS] Masculine singular ordinal capable of pronominal function
(primero, segundo)
\item[PAL   ] Portmanteau word formed by {\it a} and {\it el}
\item[PDEL  ] Portmanteau word formed by {\it de} and {\it el}
\item[PE    ] Foreign word
\item[PNC   ] Unclassified word
\item[PPC1P ] Clitic personal pronoun, first person plural DO/IO (nos)
\item[PPC1S ] Clitic personal pronoun, first person singular DO/IO (me)
\item[PPC2P ] Clitic personal pronoun, second person plural DO/IO (os)
\item[PPC2S ] Clitic personal pronoun, second person singular DO/IO (te)
\item[PPC3P ] Clitic personal pronoun, third person plural DO/IO (les)
\item[PPC3S ] Clitic personal pronoun, third person singular DO/IO (le)
\item[PPN1S ] Personal pronoun, first person singular nominative (yo)
\item[PPN2S ] Personal pronoun, second person singular nominative (t\'u)
\item[PPO3FP] Clitic personal pronoun, feminine third person plural DO (las)
\item[PPO3FS] Clitic personal pronoun, feminine third person singular DO (la)
\item[PPO3MP] Clitic personal pronoun, masculine third person plural DO (los)
\item[PPO3XS] Clitic personal pronoun, masculine or neuter third person
singular DO (lo)
\item[PPOSFP] Feminine plural possessive pronoun (tuyas, suyas)
\item[PPOSFS] Feminine singular possissive pronoun (m\'{\i}a, tuya)
\item[PPOSMP] Masculine plural possessive pronoun (m\'{\i}os, tuyos)
\item[PPOSMS] Masculine singular possessive pronoun (tuyo, suyo)
\item[PPOSPP] Plural prenominal possessive pronoun (mis, tus, sus)
\item[PPOSPS] Singular prenominal possessive pronoun (mi, tu, su)
\item[PPP1S ] Personal pronoun, first person singular oblique (m\'{\i})
\item[PPP2S ] Personal pronoun, second person singular oblique (ti)
\item[PPP3X ] Personal pronoun, third person neutral for number oblique
(s\'{\i})
\item[PPX1FP] Personal pronoun, feminine first person plural nominative or
oblique (nos\-otras)
\item[PPX1MP] Personal pronoun, masculine first person plural nominative or
oblique (nos\-otros)
\item[PPX2FP] Personal pronoun, feminine second person plural nominative or
oblique (vos\-otras)
\item[PPX2MP] Personal pronoun, masculine second person plural nominative or
oblique (vos\-otros)
\item[PPX3FP] Personal pronoun, feminine third person plural nominative or
oblique (ellas)
\item[PPX3FS] Personal pronoun, feminine third person singular nominative or
oblique (ella)
\item[PPX3MP] Personal pronoun, masculine third person plural nominative or
oblique (ellos)
\item[PPX3MS] Personal pronoun, masculine third person singular nominative or
oblique (\'el)
\item[PPX3NS] Personal pronoun, neuter third person singular nominative or
oblique (ello)
\item[PPXT2P] Personal pronoun, second person plural polite nominative or
oblique (ustedes)
\item[PPXT2S] Personal pronoun,  second person singular polite nominative or
oblique (usted)
\item[PREP  ] Preposition
\item[PREPN ] Negative preposition (sin)
\item[QUDF  ] Feminine plural distributive quantifier (sendas)
\item[QUDM  ] Masculine plural distributive quantifier (sendos)
\item[QUDX  ] Distributive quantifier neutral for gender and number (cada)
\item[QUPMUL] Singular pronominal quantifier that indicates multiples (doble,
triple)
\item[QUNFP ] Feminine plural non pronominal quantifier (diversas)
\item[QUNFS ] Feminine singular non pronominal quantifier (cualquier)
\item[QUNMP ] Masculine plural non pronominal quantifier (diversos)
\item[QUNMS ] Masculine singular non pronominal quantifier (alg\'un)
\item[QUNNMS] Masculine singular non pronominal quantifier with negative
polarity (ning\'un)
\item[QUPA  ] Singular pronominal quantifier for animates (alguien)
\item[QUPI  ] Singular pronominal quantifier for inanimates (algo)
\item[QUPNA ] Singular pronominal quantifier for animates with negative
polarity (nadie)
\item[QUPNI ] Masculine singular pronominal quantifier for inanimates with
negative polarity (nada)
\item[QUPNX ] Masculine singular pronominal quantifier with negative polarity
underspecified for animates and inanimates (ninguno)
\item[QUXFP ] Feminine plural quantifier capable of pronominal function (todas,
algunas, cualesquiera)
\item[QUXFS ] Feminine singular quantifier capable of pronominal function
(toda, alguna, cualquiera)
\item[QUXMP ] Masculine plural quantifier capable of pronominal function
(todos, algunos, cualesquiera)
\item[QUXMS ] Masculine singular quantifier capable of pronominal function
(todo, alguno, cualquiera)
\item[QUXNFP] Femenine plural quantifier capable of pronominal function with
negative polarity (ningunas)
\item[QUXNFS] Feminine singular quantifier capable of pronominal function with
negative polarity (ningu\-na)
\item[QUXNMP] Masculine plural quantifier capable of pronominal function with
negative polarity (ningu\-nos)
\item[RELPFP] Feminine plural possessive relative pronoun (cuyas)
\item[RELPFS] Feminine singular possessive relative pronoun (cuya)
\item[RELPMP] Masculine plural possessive relative pronoun (cuyos)
\item[RELPMS] Masculine singular possessive relative pronoun (cuyo)
\item[RELPXP] Pural relative pronoun for animates, neutral for gender (quienes)
\item[RELPXS] Singular relative pronoun for animates, neutral for gender
(quien)
\item[RELXFP] Feminine plural relative pronoun capable of pronominal function
(cuantas)
\item[RELXFS] Feminine singular relative pronoun capable of pronominal function
(cuanta)
\item[RELXMP] Masculine plural relative pronoun capable of pronominal function
(cuantos)
\item[RELXMS] Masculine singular relative pronoun capable of pronominal
function (cuanto)
\item[SE    ] {\it Se} (as particle)
\item[TRATF ] Feminine noun of title (Sra. D\~na. Exma.)
\item[TRATM ] Masculine noun of title (Sr., D., Prof., Exmo.)
\item[UMFX  ] Feminine unit of measurement, neutral for number (pta.)
\item[UMMX  ] Masculine unit of measurement, neutral for number (cm.)
\item[VECI1P] Verb {\it estar}. Indicative conditional tense first person
plural
\item[VECI1S] Verb {\it estar}. Indicative conditional tense first person
singular
\item[VECI2P] Verb {\it estar}. Indicative conditional tense second person
plural
\item[VECI2S] Verb {\it estar}. Indicative conditional tense second person
singular
\item[VECI3P] Verb {\it estar}. Indicative conditional tense thrid person
plural
\item[VECI3S] Verb {\it estar}. Indicative conditional tense third person
singular
\item[VEFI1P] Verb {\it estar}. Indicative future tense first person plural
\item[VEFI1S] Verb {\it estar}. Indicative future tense first person singular
\item[VEFI2P] Verb {\it estar}. Indicative future tense second person plural
\item[VEFI2S] Verb {\it estar}. Indicative future tense second person singular
\item[VEFI3P] Verb {\it estar}. Indicative future tense third person plural
\item[VEFI3S] Verb {\it estar}. Indicative future tense third person singular
\item[VEFS1P] Verb {\it estar}. Subjunctive future tense first person plural
\item[VEFS1S] Verb {\it estar}. Subjunctive future tense first person singular
\item[VEFS2P] Verb {\it estar}. Subjunctive future tense second person plural
\item[VEFS2S] Verb {\it estar}. Subjunctive future tense second person singular
\item[VEFS3P] Verb {\it estar}. Subjunctive future tense third person plural
\item[VEFS3S] Verb {\it estar}. Subjunctive future tense third person singular
\item[VEGER ] Verb {\it estar}. Gerund
\item[VEII1P] Verb {\it estar}. Indicative imperfect tense first person plural
\item[VEII1S] Verb {\it estar}. Indicative imperfect tense first person
singular
\item[VEII2P] Verb {\it estar}. Indicative imperfect tense second person plural
\item[VEII2S] Verb {\it estar}. Indicative imperfect tense second person
singular
\item[VEII3P] Verb {\it estar}. Indicative imperfect tense third person plural
\item[VEII3S] Verb {\it estar}. Indicative imperfect tense third person
singular
\item[VEINF ] Verb {\it estar}. Infinitive
\item[VEIS1P] Verb {\it estar}. Subjunctive imperfect tense first person plural
\item[VEIS1S] Verb {\it estar}. Subjunctive imperfect tense first person
singular
\item[VEIS2P] Verb {\it estar}. Subjunctive imperfect tense second person
plural
\item[VEIS2S] Verb {\it estar}. Subjunctive imperfect tense second person
singular
\item[VEIS3P] Verb {\it estar}. Subjunctive imperfect tense third person plural
\item[VEIS3S] Verb {\it estar}. Subjunctive imperfect tense third person
singular
\item[VEPI1P] Verb {\it estar}. Indicative present tense first person plural
\item[VEPI1S] Verb {\it estar}. Indicative present tense first person singular
\item[VEPI2P] Verb {\it estar}. Indicative present tense second person plural
\item[VEPI2S] Verb {\it estar}. Indicative present tense second person singular
\item[VEPI3P] Verb {\it estar}. Indicative present tense third person plural
\item[VEPI3S] Verb {\it estar}. Indicative present tense third person singular
\item[VEPM2P] Verb {\it estar}. Imperative second person plural
\item[VEPM2S] Verb {\it estar}. Imperative second person singular
\item[VEPS1P] Verb {\it estar}. Subjunctive present tense first person plural
\item[VEPS1S] Verb {\it estar}. Subjunctive present tense first person singular
\item[VEPS2P] Verb {\it estar}. Subjunctive present tense second person plural
\item[VEPS2S] Verb {\it estar}. Subjunctive present tense second person
singular
\item[VEPS3P] Verb {\it estar}. Subjunctive present tense third person plural
\item[VEPS3S] Verb {\it estar}. Subjunctive present tense third person singular
\item[VEPX  ] Verb {\it estar}. Past participle
\item[VEXI1P] Verb {\it estar}. Indicative preterite tense first person plural
\item[VEXI1S] Verb {\it estar}. Indicative preterite tense first person
singular
\item[VEXI2P] Verb {\it estar}. Indicative preterite tense second person plural
\item[VEXI2S] Verb {\it estar}. Indicative preterite tense second person
singular
\item[VEXI3P] Verb {\it estar}. Indicative preterite tense third person plural
\item[VEXI3S] Verb {\it estar}. Indicative preterite tense third person
singular
\item[VHCI1P] Verb {\it haber}. Indicative conditional tense first person
plural
\item[VHCI1S] Verb {\it haber}. Indicative conditional tense first person
singular
\item[VHCI2P] Verb {\it haber}. Indicative conditional tense second person
plural
\item[VHCI2S] Verb {\it haber}. Indicative conditional tense second person
singular
\item[VHCI3P] Verb {\it haber}. Indicative conditional tense third person
plural
\item[VHCI3S] Verb {\it haber}. Indicative conditional tense third person
singular
\item[VHFI1P] Verb {\it haber}. Indicative future tense first person plural
\item[VHFI1S] Verb {\it haber}. Indicative future tense first person singular
\item[VHFI2P] Verb {\it haber}. Indicative future tense second person plural
\item[VHFI2S] Verb {\it haber}. Indicative future tense second person singular
\item[VHFI3P] Verb {\it haber}. Indicative future tense third person plural
\item[VHFI3S] Verb {\it haber}. Indicative future tense thrid person singular
\item[VHFS1P] Verb {\it haber}. Subjunctive future tense first person plural
\item[VHFS1S] Verb {\it haber}. Subjunctive future tense first person singular
\item[VHFS2P] Verb {\it haber}. Subjunctive future tense second person plural
\item[VHFS2S] Verb {\it haber}. Subjunctive future tense second person singular
\item[VHFS3P] Verb {\it haber}. Subjunctive future tense third person plural
\item[VHFS3S] Verb {\it haber}. Subjunctive future tense third person singular
\item[VHGER ] Verb {\it haber}. Gerund
\item[VHII1P] Verb {\it haber}. Indicative imperfect tense first person plural
\item[VHII1S] Verb {\it haber}. Indicative imperfect tense first person
singular
\item[VHII2P] Verb {\it haber}. Indicative imperfect tense second person plural
\item[VHII2S] Verb {\it haber}. Indicative imperfect tense second person
singular
\item[VHII3P] Verb {\it haber}. Indicative imperfect tense third person plural
\item[VHII3S] Verb {\it haber}. Indicative imperfect tense third person
singular
\item[VHINF ] Verb {\it haber}. Infinitive
\item[VHIS1P] Verb {\it haber}. Subjunctive imperfect tense first person plural
\item[VHIS1S] Verb {\it haber}. Subjunctive imperfect tense first person
singular
\item[VHIS2P] Verb {\it haber}. Subjunctive imperfect tense second person
plural
\item[VHIS2S] Verb {\it haber}. Subjunctive imperfect tense second person
singular
\item[VHIS3P] Verb {\it haber}. Subjunctive imperfect tense third person plural
\item[VHIS3S] Verb {\it haber}. Subjunctive imperfect tense third person
singular
\item[VHPI1P] Verb {\it haber}. Indicative present tense first person plural
\item[VHPI1S] Verb {\it haber}. Indicative present tense first person singular
\item[VHPI2P] Verb {\it haber}. Indicative present tense second person plural
\item[VHPI2S] Verb {\it haber}. Indicative present tense second person singular
\item[VHPI3E] Verb {\it haber}. Indicative present tense third person singular
existential
\item[VHPI3P] Verb {\it haber}. Indicative present tense third person plural
\item[VHPI3S] Verb {\it haber}. Indicative present tense third person singular
\item[VHPS1P] Verb {\it haber}. Subjunctive present tense first person plural
\item[VHPS1S] Verb {\it haber}. Subjunctive present tense first person singular
\item[VHPS2P] Verb {\it haber}. Subjunctive present tense second person plural
\item[VHPS2S] Verb {\it haber}. Subjunctive present tense second person
singular
\item[VHPS3P] Verb {\it haber}. Subjunctive present tense third person plural
\item[VHPS3S] Verb {\it haber}. Subjunctive present tense third person singular
\item[VHPXFP] Verb {\it haber}. Feminine plural past participle
\item[VHPXFS] Verb {\it haber}. Feminine singular past participle
\item[VHPXMP] Verb {\it haber}. Masculine plural past participle
\item[VHPXMS] Verb {\it haber}. Masculine singular past participle
\item[VHXI1P] Verb {\it haber}. Indicative preterite tense first person plural
\item[VHXI1S] Verb {\it haber}. Indicative preterite tense first person
singular
\item[VHXI2P] Verb {\it haber}. Indicative preterite tense second person plural
\item[VHXI2S] Verb {\it haber}. Indicative preterite tense second person
singular
\item[VHXI3P] Verb {\it haber}. Indicative preterite tense third person plural
\item[VHXI3S] Verb {\it haber}. Indicative preterite tense third person
singular
\item[VLCI1P] Lexical verb. Indicative conditional tense first person plural
\item[VLCI1S] Lexical verb. Indicative conditional tense first person singular
\item[VLCI2P] Lexical verb. Indicative conditional tense second person plural
\item[VLCI2S] Lexical verb. Indicative conditional tense second person singular
\item[VLCI3P] Lexical verb. Indicative conditional tense third person plural
\item[VLCI3S] Lexical verb. Indicative conditional tense third person singular
\item[VLFI1P] Lexical verb. Indicative future tense first person plural
\item[VLFI1S] Lexical verb. Indicative future tense first person singular
\item[VLFI2P] Lexical verb. Indicative future tense second person plural
\item[VLFI2S] Lexical verb. Indicative future tense second person singular
\item[VLFI3P] Lexical verb. Indicative future tense third person plural
\item[VLFI3S] Lexical verb. Indicative future tense thrid person singular
\item[VLFS1P] Lexical verb. Subjunctive future tense first person plural
\item[VLFS1S] Lexical verb. Subjunctive future tense first person singular
\item[VLFS2P] Lexical verb. Subjunctive future tense second person plural
\item[VLFS2S] Lexical verb. Subjunctive future tense second person singular
\item[VLFS3P] Lexical verb. Subjunctive future tense third person plural
\item[VLFS3S] Lexical verb. Subjunctive future tense third person singular
\item[VLGER ] Lexical verb. Gerund
\item[VLII1P] Lexical verb. Indicative imperfect tense first person plural
\item[VLII1S] Lexical verb. Indicative imperfect tense first person singular
\item[VLII2P] Lexical verb. Indicative imperfect tense second person plural
\item[VLII2S] Lexical verb. Indicative imperfect tense second person singular
\item[VLII3P] Lexical verb. Indicative imperfect tense third person plural
\item[VLII3S] Lexical verb. Indicative imperfect tense third person singular
\item[VLINF ] Lexical verb. Infinitive
\item[VLIS1P] Lexical verb. Subjunctive imperfect tense first person plural
\item[VLIS1S] Lexical verb. Subjunctive imperfect tense first person singular
\item[VLIS2P] Lexical verb. Subjunctive imperfect tense second person plural
\item[VLIS2S] Lexical verb. Subjunctive imperfect tense second person singular
\item[VLIS3P] Lexical verb. Subjunctive imperfect tense third person plural
\item[VLIS3S] Lexical verb. Subjunctive imperfect tense third person singular
\item[VLPI1P] Lexical verb. Indicative present tense first person plural
\item[VLPI1S] Lexical verb. Indicative present tense first person singular
\item[VLPI2P] Lexical verb. Indicative present tense second person plural
\item[VLPI2S] Lexical verb. Indicative present tense second person singular
\item[VLPI3P] Lexical verb. Indicative present tense third person plural
\item[VLPI3S] Lexical verb. Indicative present tense third person singular
\item[VLPM2P] Lexical verb. Imperative second person plural
\item[VLPM2S] Lexical verb. Imperative second person singular
\item[VLPPFP] Lexical verb. Feminine plural present participle
\item[VLPPFS] Lexical verb. Feminine singular present participle
\item[VLPPMP] Lexical verb. Masculine plural present participle
\item[VLPPMS] Lexical verb. Masculine singular present participle
\item[VLPS1P] Lexical verb. Subjunctive present tense first person plural
\item[VLPS1S] Lexical verb. Subjunctive present tense first person singular
\item[VLPS2P] Lexical verb. Subjunctive present tense second person plural
\item[VLPS2S] Lexical verb. Subjunctive present tense second person singular
\item[VLPS3P] Lexical verb. Subjunctive present tense third person plural
\item[VLPS3S] Lexical verb. Subjunctive present tense third person singular
\item[VLPXFP] Lexical verb. Feminine plural past participle
\item[VLPXFS] Lexical verb. Feminine singular past participle
\item[VLPXMP] Lexical verb. Masculine plural past participle
\item[VLPXMS] Lexical verb. Masculine singular past participle
\item[VLXI1P] Lexical verb. Indicative preterite tense first person plural
\item[VLXI1S] Lexical verb. Indicative preterite tense first person singular
\item[VLXI2P] Lexical verb. Indicative preterite tense second person plural
\item[VLXI2S] Lexical verb. Indicative preterite tense second person singular
\item[VLXI3P] Lexical verb. Indicative preterite tense third person plural
\item[VLXI3S] Lexical verb. Indicative preterite tense third person singular
\item[VMCI1P] Modal verb. Indicative conditional tense first person plural
\item[VMCI1S] Modal verb. Indicative conditional tense first person singular
\item[VMCI2P] Modal verb. Indicative conditional tense second person plural
\item[VMCI2S] Modal verb. Indicative conditional tense second person singular
\item[VMCI3P] Modal verb. Indicative conditional tense third person plural
\item[VMCI3S] Modal verb. Indicative conditional tense third person singular
\item[VMFI1P] Modal verb. Indicative future tense first person plural
\item[VMFI1S] Modal verb. Indicative future tense first person singular
\item[VMFI2P] Modal verb. Indicative future tense second person plural
\item[VMFI2S] Modal verb. Indicative future tense second person singular
\item[VMFI3P] Modal verb. Indicative future tense third person plural
\item[VMFI3S] Modal verb. Indicative future tense thrid person singular
\item[VMFS1P] Modal verb. Subjunctive future tense first person plural
\item[VMFS1S] Modal verb. Subjunctive future tense first person singular
\item[VMFS2P] Modal verb. Subjunctive future tense second person plural
\item[VMFS2S] Modal verb. Subjunctive future tense second person singular
\item[VMFS3P] Modal verb. Subjunctive future tense third person plural
\item[VMFS3S] Modal verb. Subjunctive future tense third person singular
\item[VMGER ] Modal verb. Gerund
\item[VMII1P] Modal verb. Indicative imperfect tense first person plural
\item[VMII1S] Modal verb. Indicative imperfect tense first person singular
\item[VMII2P] Modal verb. Indicative imperfect tense second person plural
\item[VMII2S] Modal verb. Indicative imperfect tense second person singular
\item[VMII3P] Modal verb. Indicative imperfect tense third person plural
\item[VMII3S] Modal verb. Indicative imperfect tense third person singular
\item[VMINF ] Modal verb. Infinitive
\item[VMIS1P] Modal verb. Subjunctive imperfect tense first person plural
\item[VMIS1S] Modal verb. Subjunctive imperfect tense first person singular
\item[VMIS2P] Modal verb. Subjunctive imperfect tense second person plural
\item[VMIS2S] Modal verb. Subjunctive imperfect tense second person singular
\item[VMIS3P] Modal verb. Subjunctive imperfect tense third person plural
\item[VMIS3S] Modal verb. Subjunctive imperfect tense third person singular
\item[VMPI1P] Modal verb. Indicative present tense first person plural
\item[VMPI1S] Modal verb. Indicative present tense first person singular
\item[VMPI2P] Modal verb. Indicative present tense second person plural
\item[VMPI2S] Modal verb. Indicative present tense second person singular
\item[VMPI3P] Modal verb. Indicative present tense third person plural
\item[VMPI3S] Modal verb. Indicative present tense third person singular
\item[VMPM2P] Modal verb. Imperative second person plural
\item[VMPM2S] Modal verb. Imperative second person singular
\item[VMPS1P] Modal verb. Subjunctive present tense first person plural
\item[VMPS1S] Modal verb. Subjunctive present tense first person singular
\item[VMPS2P] Modal verb. Subjunctive present tense second person plural
\item[VMPS2S] Modal verb. Subjunctive present tense second person singular
\item[VMPS3P] Modal verb. Subjunctive present tense third person plural
\item[VMPS3S] Modal verb. Subjunctive present tense third person singular
\item[VMPX  ] Modal verb. Past participle
\item[VMXI1P] Modal verb. Indicative preterite tense first person plural
\item[VMXI1S] Modal verb. Indicative preterite tense first person singular
\item[VMXI2P] Modal verb. Indicative preterite tense second person plural
\item[VMXI2S] Modal verb. Indicative preterite tense second person singular
\item[VMXI3P] Modal verb. Indicative preterite tense third person plural
\item[VMXI3S] Modal verb. Indicative preterite tense third person singular
\item[VSCI1P] Verb {\it ser}. Indicative conditional tense first person plural
\item[VSCI1S] Verb {\it ser}. Indicative conditional tense first person
singular
\item[VSCI2P] Verb {\it ser}. Indicative conditional tense second person plural
\item[VSCI2S] Verb {\it ser}. Indicative conditional tense second person
singular
\item[VSCI3P] Verb {\it ser}. Indicative conditional tense third person plural
\item[VSCI3S] Verb {\it ser}. Indicative conditional tense third person
singular
\item[VSFI1P] Verb {\it ser}. Indicative future tense first person plural
\item[VSFI1S] Verb {\it ser}. Indicative future tense first person singular
\item[VSFI2P] Verb {\it ser}. Indicative future tense second person plural
\item[VSFI2S] Verb {\it ser}. Indicative future tense second person singular
\item[VSFI3P] Verb {\it ser}. Indicative future tense third person plural
\item[VSFI3S] Verb {\it ser}. Indicative future tense thrid person singular
\item[VSFS1P] Verb {\it ser}. Subjunctive future tense first person plural
\item[VSFS1S] Verb {\it ser}. Subjunctive future tense first person singular
\item[VSFS2P] Verb {\it ser}. Subjunctive future tense second person plural
\item[VSFS2S] Verb {\it ser}. Subjunctive future tense second person singular
\item[VSFS3P] Verb {\it ser}. Subjunctive future tense third person plural
\item[VSFS3S] Verb {\it ser}. Subjunctive future tense third person singular
\item[VSGER ] Verb {\it ser}. Gerund
\item[VSII1P] Verb {\it ser}. Indicative imperfect tense first person plural
\item[VSII1S] Verb {\it ser}. Indicative imperfect tense first person singular
\item[VSII2P] Verb {\it ser}. Indicative imperfect tense second person plural
\item[VSII2S] Verb {\it ser}. Indicative imperfect tense second person singular
\item[VSII3P] Verb {\it ser}. Indicative imperfect tense third person plural
\item[VSII3S] Verb {\it ser}. Indicative imperfect tense third person singular
\item[VSINF ] Verb {\it ser}. Infinitive
\item[VSIS1P] Verb {\it ser}. Subjunctive imperfect tense first person plural
\item[VSIS1S] Verb {\it ser}. Subjunctive imperfect tense first person singular
\item[VSIS2P] Verb {\it ser}. Subjunctive imperfect tense second person plural
\item[VSIS2S] Verb {\it ser}. Subjunctive imperfect tense second person
singular
\item[VSIS3P] Verb {\it ser}. Subjunctive imperfect tense third person plural
\item[VSIS3S] Verb {\it ser}. Subjunctive imperfect tense third person singular
\item[VSPI1P] Verb {\it ser}. Indicative present tense first person plural
\item[VSPI1S] Verb {\it ser}. Indicative present tense first person singular
\item[VSPI2P] Verb {\it ser}. Indicative present tense second person plural
\item[VSPI2S] Verb {\it ser}. Indicative present tense second person singular
\item[VSPI3P] Verb {\it ser}. Indicative present tense third person plural
\item[VSPI3S] Verb {\it ser}. Indicative present tense third person singular
\item[VSPM2P] Verb {\it ser}. Imperative second person plural
\item[VSPM2S] Verb {\it ser}. Imperative second person singular
\item[VSPS1P] Verb {\it ser}. Subjunctive present tense first person plural
\item[VSPS1S] Verb {\it ser}. Subjunctive present tense first person singular
\item[VSPS2P] Verb {\it ser}. Subjunctive present tense second person plural
\item[VSPS2S] Verb {\it ser}. Subjunctive present tense second person singular
\item[VSPS3P] Verb {\it ser}. Subjunctive present tense third person plural
\item[VSPS3S] Verb {\it ser}. Subjunctive present tense third person singular
\item[VSPX  ] Verb {\it ser}. Past participle
\item[VSXI1P] Verb {\it ser}. Indicative preterite tense first person plural
\item[VSXI1S] Verb {\it ser}. Indicative preterite tense first person singular
\item[VSXI2P] Verb {\it ser}. Indicative preterite tense second person plural
\item[VSXI2S] Verb {\it ser}. Indicative preterite tense second person singular
\item[VSXI3P] Verb {\it ser}. Indicative preterite tense third person plural
\item[VSXI3S] Verb {\it ser}. Indicative preterite tense third person singular

\end{description}

\end{document}